\documentclass{article}

\usepackage{verbatim}
\usepackage{amsmath}
\usepackage{amssymb}
\usepackage{natbib}

\newcommand\E{\mathop{\mbox{{\rm E}}}\limits}
\usepackage{graphicx}
\usepackage{subfig}
\usepackage{algorithm}
\usepackage{algorithmic}

\bibpunct{(}{)}{;}{a}{,}{,}

\title{A Hybrid Method for Distance Metric Learning}
\author{
Yi-Hao Kao \footnote{Corresponding author contact: yhkao@alumni.stanford.edu} \\
\hspace{1.5in}
\and
Benjamin Van Roy \\
\hspace{1.5in}
\and
Daniel Rubin \\
\hspace{1.5in}
\and
Jiajing Xu \\
\hspace{1.5in}
\and
Jessica Faruque \\
\\
\emph{Stanford University}
\and
Sandy Napel \\
\hspace{1.5in}
}

\date{May 26, 2011}

\begin{document}

\maketitle

\begin{abstract} 
We consider the problem of learning a measure of distance among vectors in a feature space 
and propose a hybrid method that simultaneously learns from similarity ratings assigned to pairs of vectors 
and class labels assigned to individual vectors.  Our method is based on a generative model in which
class labels can provide information that is not encoded in feature vectors but yet relates to 
perceived similarity between objects.
Experiments with synthetic data as well as a real medical image retrieval problem demonstrate that leveraging class labels through
use of our method improves retrieval performance significantly.
\end{abstract} 

\section{Introduction}

Consider a retrieval system that, given features of an object, searches a database for similar objects.  Such a system requires a distance metric for assessing similarity.  One way to produce a distance metric is to learn from similarity ratings that representative users have assigned to pairs of objects.  Given data of this kind, ratings can be regressed onto differences between object features.

In this paper, we consider the use of class labels in addition to similarity ratings to learn a distance metric. Labels may be available, for example, if each object is assigned a class when entered into the database.  The class label does not serve as an additional feature because when searching for objects similar to a new one, the class of the new object is usually unknown. In fact, the purpose of the retrieval system may be to supply similar objects and their class labels to assist the user in classifying the new object. However, class labels provide information useful to learning the distance metric because they may relate to similarity ratings in ways not captured by extracted features.

While distance metric learning has attracted much attention in recent years, approaches that have been proposed generally learn from either similarity/difference data or class labels but not both. We will refer to these two types of approaches as similarity-based and class-based methods, respectively. In the former category are multidimensional scaling methods \citep{Cox00}, which embed vectors in a Euclidean space so that distances between pairs are close to available estimates, ordinal regression \citep{McCullagh89, Herbrich00}, which learns a function that maps feature differences to discrete levels of measured similarity, and convex optimization formulations \citep{Xing02,Schultz04,Frome06}, which learn metrics that tend to make data pairs classified as similar close and others distant.  As for class-based methods, examples include relevant component analysis \citep{Bar-Hillel03}, which aims to learn a metric that makes data points that share a class close and others distant, neighbourhood component analysis \citep{Goldberger05}, which learns a distance metric by optimizing the probability of correct classification based on a softmax model and nearest neighbors, and the algorithms of \citet{Weinberger06},\citet{Weinberger07}, and \citet{Weinberger09}, which minimize the distances between objects in each neighborhood that share the same class while separating those from different classes.

Our hybrid method of distance metric learning advances the aforementioned literature
by providing an effective algorithm that makes use of both kinds of data simultaneously. It consists of two stages: a soft classifier is learned from the class label data
and then used together with the similarity rating data by any similarity-based distance metric learning algorithm.
Although this method can make use of any algorithm for learning a soft classifier and any similarity-based distance metric learning algorithm, to best illustrate our idea we will focus on the combination of a kernel density estimation algorithm similar to neighborhood component analysis and the aforementioned convex optimization approach to learning from similarity ratings.
Results from experiments with synthetic data as well as a real medical image retrieval problem demonstrate that this hybrid method improves retrieval performance significantly.

\section{Problem Formulation}
\subsection{Data}
Suppose features of each object are encoded in a vector $x \in \mathbb{R}^K$.  We are given a data set consisting of similarity ratings for pairs of objects and class labels for individual objects.  The ratings data is comprised of a set $\cal S$ of quintuplets $(o, o', x,x',\sigma)$,
each consisting of two object identifiers $o$ and $o'$, associated feature vectors $x$ and $x'$, and a similarity rating $\sigma$.  We assume that each similarity  rating 
takes one of three values, in particular, $1$, $2$, and $3$,  conveying dissimilarity, neutrality, and similarity, respectively. Denote the number of classes by $M$ and 
index each class by an integer from $1$ through $M$. The class label data is a set $\cal G$ of triplets $(o,x,c)$, each consisting of an
object identifier $o$, a feature vector $x$, and a class $c \in \{1,2,\ldots, M \}$.  The reason that object identifiers are included in the data is so that we know when a given class label is associated
with the same object as a given similarity rating.  In order to compress notation, when the object identifiers are not relevant to a discussion, we will refer to data samples
in $\cal S$ as triplets $(x,x',\sigma)$ and data in $\cal G$ as pairs $(x,c)$.

\subsection{Distance Metric}
A distance metric is a mapping from $\mathbb{R}^K \times \mathbb{R}^K$ to $\mathbb{R}_+$ which assesses the distance of any given pair of objects.  Given a
a class of distance metrics $d_r: \mathbb{R}^K \times \mathbb{R}^K \rightarrow \mathbb{R}_+$, which is parameterized by a vector $r$, we wish to compute $r$ so that the resulting distance metric accurately reflects perceived distances.  Though the methods we present apply to a variety of distance metrics, much of our discussion will focus on the popular choice of
a weighted Euclidean norm:
\begin{equation}
d_r(x, x') = \sqrt{ \sum_{k=1}^K r_k (x_k - x'_k )^2 } . \label{eq:d_r}
\end{equation}

\section{Algorithms}
Our goal is to learn a distance metric $d: \mathbb{R}^K \times \mathbb{R}^K \rightarrow \mathbb{R}_+$ that help us retrieve similar objects in the database. We now discuss three existing algorithms for doing so and propose a new hybrid algorithm.

\subsection{Ordinal Regression}

Ordinal regression \citep{McCullagh89} offers a simple approach to learning coefficients from the similarity rating data $\cal S$.
Ordinal regression typically assumes that given a pair of objects $(x,x')$, similarity ratings obeys the conditional distribution
$$P( \sigma \leq v | x,x' ) = \frac{1}{1+\exp( -d_r(x,x')^2-\theta_v ) }$$
where $v\in \{1,2,3 \}$ denotes the level of similarity, and $\theta_1 \leq \theta_2$ are boundary parameters (we have implicitly $\theta_3=\infty$ ). These parameters, together with the coefficients $r$, are computed by
solving a maximum likelihood problem:
\begin{eqnarray*}
\max_{r, \theta} && \sum\limits_{(x, x',\sigma)\in {\cal S}} \log P(\sigma | x, x') \\
\mbox{s.t } && r \geq 0\\
&& \theta_1 \leq \theta_2.
\end{eqnarray*}
Constraints are imposed on $r$ because,
given the way our distance metric is defined in (\ref{eq:d_r}), coefficients of any suitable distance metric should be nonnegative. Note that this algorithm only makes use of 
the rating data $\cal S$.

\subsection{Convex Optimization}
\label{sec:cvx_opt}

Another approach, proposed in \citet{Xing02}, computes $r$ by solving a convex optimization problem:
\begin{eqnarray*}
\min_r && \sum\limits_{(x, x',\sigma=3)\in {\cal S}}  d^2_r(x, x') \\
\mbox{s.t.} && \sum\limits_{ (x, x',\sigma=1)\in {\cal S}}  d_r(x, x') \geq 1 \\
&& r\geq 0.
\end{eqnarray*}
This formulation results in a distance metric that aims to minimize the distances between similar objects while keeping dissimilar ones sufficiently far apart.  Similarly with ordinal regression, this algorithm only makes use of the rating data $\cal S$.

\subsection{Neighborhood Component Analysis}

Neighborhood component analysis (NCA) learns a distance metric from class labels based on an assumption that similar objects are more likely to share the same
class than dissimilar ones.  NCA employs a model in which a feature vector $x^\dagger$  is assigned class label 
$c^\dagger$ with probability
\begin{equation}
 P(c^\dagger | x^\dagger, {\cal G}) = \frac{ \sum\limits_{(x,c=c^\dagger) \in {\cal G}} \exp(-d^2_r(x^\dagger, x )) }{ \sum\limits_{(x',c') \in {\cal G}} \exp( -d^2_r(x^\dagger, x'))}.
\label{eq:NCA_P}
\end{equation}
NCA computes coefficients that would lead to accurate classification of objects in the training set $\cal G$.  We will define accuracy here in terms of log likelihood.  
In particular, we consider an implementation that aims to produce coefficients by maximizing the average leave-one-out log-likelihood. That is,
\begin{equation}
\max_{r\geq 0} \sum_{(x,c) \in {\cal G}} \log P\big(c | x, {\cal G} \setminus{(x,c)}\big). \label{eq:NCA_objective}
\end{equation}
This optimization problem is not convex, but in our experience a local-optimum can be found efficiently via projected gradient ascent. In many practical cases the number of training samples is not much larger than the number of parameters $K$, and NCA consequently suffers from overfitting.  Therefore, we consider $L_1$ regularization in our application of NCA.  In particular, we subtract a penalty term $\lambda \|r\|_1$ from (\ref{eq:NCA_objective}), where the parameter $\lambda$ is selected by cross-validation.  Further details about our implementation can be found in the appendix.

\subsection{A Hybrid Method}
We now introduce a hybrid method that simultaneously makes use of similarity ratings and class labels. Our approach is motivated by an assumption that similarity ratings are driven by a weighted Euclidean norm distance metric, but that the observed
feature vectors may not express all relevant information about objects being compared.  In particular, there may be ``missing features'' that influence the underlying
distance metric.  Given objects $o$ and $o'$ with observed feature vectors $x, x' \in \mathbb{R}^K$ and missing feature vectors $z, z' \in \mathbb{R}^J$, we assume
the underlying distance metric is given by
\begin{eqnarray*}
{\cal D}(o,o') 
&=& \left(\sum_{k=1}^K r_k (x_k - x'_k)^2 + \sum_{j=1}^J r^\perp_j (z_j - z'_j)^2\right)^\frac{1}{2} \\
&=& \left(d^2_r(x,x') + d^2_{r^\perp}(z, z')\right)^{1/2},
\end{eqnarray*}
where $r \in \mathbb{R}_+^K$ and $r^\perp \in \mathbb{R}_+^J$.

Another important assumption we will make concerning the missing feature vector is that it is conditionally independent
from the observed feature vector when conditioned on the class label.  In other words, given an object with observed and missing feature vectors $x$ and $z$ 
and a class label $c$, we have $p(x,z|c) = p(x|c) p(z|c)$. This assumption is justifiable since, if there exists any correlation between $x$ and $z$, then we can subtract this dependence from $z$, resulting in another random variable $z'$, and replace $z$ by $z'$ without loss of generality.

Now suppose we are given a learning algorithm $ {\cal A} $ that learns the conditional class probabilities $P(c|x)$ from class data $\cal G$. In other words, ${\cal A}$ is a function that maps $\cal G$ into an estimate $\hat{P}(\cdot|\cdot)$. Using these conditional class probabilities $\hat{P}$, we generate a soft class label for each unlabeled object represented in ${\cal S}$, our similarity ratings data set, that is not labeled in the class data set $\cal G$.  
In particular, for an unlabeled object $o$ with feature vector $x$, we generate a vector $u(o) \in \mathbb{R}^M$, with each $m$th component
given by $u_m(o) = \hat{P}(m|x)$.  For uniformity of notation, we also define for each object $o$ from ${\cal G}$, the set with class labels, 
a vector $u(o)$.  In this case, if $c$ is the class label assigned to $o$ then $u_c(o) = 1$ and $u_m(o) = 0$ for $m \neq c$.

We now discuss how the similarity ratings data ${\cal S}$ is used together with these 
class probability vectors to produce a distance metric.  The main idea is to generate an estimate of $\left(\E[{\cal D}^2(o,o') | x,x',u(o),u(o')]\right)^\frac{1}{2}$ that is consistent with observed similarity ratings. The conditioning on $u(x)$ and $u(x')$ here indicates that these vectors are taken to be the class probabilities associated with the two objects.

Note that
\begin{eqnarray*}
&& \E[{\cal D}^2(o,o') | x, x', u(o), u(o')] \\
&=& d^2_r(x,x') + \E[d^2_{r^\perp}(z, z') | x,x', u(o), u(o')]
\end{eqnarray*}
and using the conditional independence assumption we have
\begin{eqnarray*}
&& \E[d^2_{r^\perp}(z, z') | x,x', u(o), u(o')] \\
&=& \sum_{c,c'} \E[d^2_{r^\perp}(z, z') | x,x', c, c'] u_c(o) u_{c'}(o') \\
&=& \sum_{c,c'} \E[d^2_{r^\perp}(z, z') | c, c'] u_c(o) u_{c'}(o') \\
&=& u(o)^\top Q u(o'),
\end{eqnarray*}
where $Q \in \mathbb{R}^{M\times M}$ is defined as
$$ Q_{c,c'} = \E[d^2_{r^\perp}(z, z') | c, c'], \quad 1\leq c,c' \leq M. $$
We can view $Q$ as a matrix that encodes distance information relating to missing features. This motivates the following parameterization of a distance metric, which is what we will use:
\begin{eqnarray*}
d^h_{r,Q}(o,o') &=& \left(\E[{\cal D}^2(o,o') | x, x', u(o), u(o')]\right)^\frac{1}{2} \\
&=& \left(d^2_r(x,x') + u(o)^\top Q u(o')\right)^\frac{1}{2}.
\end{eqnarray*}
Note that in the event that class labels are not provided for $o$ and $o'$, the class probability vectors depend only on $x$ and $x'$.  Therefore, with
some abuse of notation, when there are no class labels, we can write the distance metric as
$$d^h_{r,Q}(x,x') = \left(d^2_r(x,x') + u(x)^\top Q u(x')\right)^\frac{1}{2}.$$

Our hybrid method estimates the vector $r \in \mathbb{R}^K$ and matrix $Q \in \mathbb{R}^{M\times M}$ so that they are consistent with similarity ratings.  To do so, it makes use of a similarity-based learning algorithm $\cal B$ that learns the coefficients of a distance metric from feature differences and similarity ratings, such as the ordinal regression or convex optimization methods we have described.

To provide a concrete version of our hybrid method, we consider the case where $\cal A$ is a kernel density estimation procedure similar to NCA and
$\cal B$ is the algorithm based on convex optimization, discussed in Section \ref{sec:cvx_opt}.  In this case, the method first generates a feature vector 
density for each class according to
$$\hat{p}(x|c) = \frac{1}{|(x',c'=c) \in {\cal G}|} \sum_{(x',c'=c) \in {\cal G}}  {\cal N}_w(x-x'),$$
where ${\cal N}_w$ is a Gaussian kernel, defined by
$${\cal N}_w(x) \propto \exp\left(-\sum_{k=1}^K w_k x_k^2\right).$$
To produce conditional class probabilities, we estimate the marginal distribution of classes according to
$$\hat{P}(c) = \frac{|(x',c'=c) \in {\cal G}|}{|{\cal G}|},$$
and applying Bayes' rule to arrive at
$$\hat{P}(c|x) = \frac{\hat{P}(c) \hat{p}(x|c)}{\sum_{m=1}^M \hat{P}(m) \hat{p}(x|m)}.$$
The Gaussian kernel parameters $w$ can be estimated by a similar approach as described in (\ref{eq:NCA_objective}). Then, to compute estimates $\hat{r}$ and $\hat{Q}$, we solve the following convex optimization problem:
\begin{eqnarray*}
\min_{r,Q} && \sum\limits_{(o,o',x, x',\sigma=3)\in {\cal S}}  d_r(x, x')^2 + u(o)^\top Qu(o') \\
\mbox{s.t.} && \sum\limits_{(o,o',x, x',\sigma=1)\in {\cal S}}  \sqrt{ d_r(x, x')^2 + u(o)^\top Qu(o') } \geq 1 \\
&& r\geq 0 \\
&& Q \geq 0 \mbox{ and symmetric.}
\end{eqnarray*}
This is the hybrid method we use in our experiments.  Note that we only require $Q$ to be element-wise non-negative, but not positive semidefinite, and as such our method does not entail solution to an SDP.

\section{Experiments}

We evaluate the aforementioned four algorithms, namely ordinal regression (OR), convex optimization (CO), neighborhood component analysis (NCA), and the hybrid method (HYB), in two experiments. In the first experiment, we generate 100 synthetic data sets by a sampling process. For the second experiment, a real data set consisting of feature vectors derived from computed tomography (CT) scans of liver lesions, along with diagnoses and comparison ratings provided by radiologists, is considered. The data was collected as part of a project that seeks to develop a similarity-based image retrieval system for radiological decision support \citep{Napel10}. We now describe the settings and empirical results of both experiments in detail.

It is worth mentioning that relative to other algorithms we consider, the hybrid method increases the number of free variables by $M(M+1) / 2$, which is the number of numerical values used to represent the symmetric matrix $Q$.  Since the number of classes $M$ is usually much smaller than the number of features $K$, we do not expect this increase in degrees of freedom 
to drive differences in empirical results.  For instance, in the medical image dataset we study, we have $K=60$ and $M=3$, so our hybrid method only introduces $6$ new variables to the $60$ variables used by other methods.

\subsection{Synthetic Data}

\begin{figure}[h]
\centering
\includegraphics[scale=0.8]{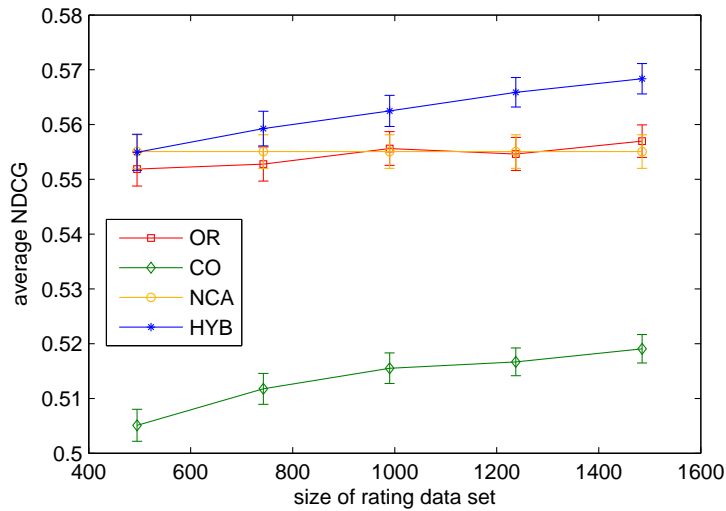}
\caption{The average NDCG$_{10}$ delivered by OR, CO, NCA, and HYB, over different sizes of rating data set. For statistical interpretation, we also give the error bars (one standard deviation) in the plots.}
\label{fig:syn_NDCG}
\end{figure}

The following procedure explains how we generate and conduct experiments with synthetic data:
\begin{enumerate}
	\item Sample a generative model and coefficient vectors $r$ and $r^\perp$. Further details about this sampling process can be found in the appendix.
	\item Generate $200$ data points from the resulting generative model; denote it by a set ${\cal O} =\{ (o^{(n)}, x^{(n)}, z^{(n)}, c^{(n)}) : n=1,2,\cdots, 200\}$.
	\item For each integer pair $(a,b), 1 \leq a,b \leq 200, a \neq b$, let
$$y^{(a,b)} =   \sum_{k=1}^K r_k|x_k^{(a)}-x_k^{(b)}|^2 + \sum_{j=1}^J r_j^\perp |z_j^{(a)}-z_j^{(b)}|^2 + \epsilon^{(a,b)} $$
	where $\epsilon^{(a,b)}$ is sampled iid from ${\cal N}(0,50^2)$ to represent the random noise in rating. This results in $39,800$ distance values. Let $y_{20\%}$ be their first quintile and $y_{50\%}$ be their median. We set
$$ \sigma^{(a,b)} = \left\{ \begin{array}{rl}
3 & \mbox{if } y^{(a,b)} < y_{20\%} \\
2 & \mbox{if } y_{20\%} \leq y^{(a,b)} < y_{50\%} \\
1 & \mbox{otherwise.} \\
\end{array} \right.$$
	\item Let ${\cal X }= \{(o^{(i)},x^{(i)}) : 1\leq i \leq 100\}$ be the training set and $\bar{\cal X } =  \{(o^{(i)},x^{(i)}) : 101 \leq i \leq 200\}$ be the testing set. Take ${\cal G} = \{ ( o^{(i)}, x^{(i)}, c^{(i)} ) : 1 \leq i \leq 100 \}$ be the label data set.
	\item Let ${\cal S} = \{ (o^{(i)}, o^{(j)},x^{(i)},x^{(j)}, \sigma^{(i,j)}) : 1\leq i , j \leq 100, i \neq j\}$ and $ \bar{{\cal S}} = \{ (o^{(i)}, o^{(j)},x^{(i)},x^{(j)}, \sigma^{(i,j)}) : 1\leq j\leq 100 < i \leq 200 \}$. $\bar{\cal S}$ will be used for testing, and for training we sample 5 subsets of ${\cal S}$, namely ${\cal S}_1, \ldots, {\cal S}_5$, such that the sizes of these sets equal to $5\%, 7.5\%, 10\%, 12.5\%$ and $15\%$ of the size of ${\cal S}$, respectively. The reason for using ${\cal S}_1, \ldots, {\cal S}_5$ as our training sets is that in many practical contexts it is not feasible to gather an exhaustive set of comparison data that rates all pairs of feature vectors as does ${\cal S}$.
	\item For $f=1,2,\ldots, 5$, run OR, CO, NCA, and HYB on the datasets $({\cal X}, {\cal G}, {\cal S}_f)$, resulting in four distance measures. Then for every $x^{(n)}\in \bar{\cal X}$, apply each distance measure to retrieve the top 10 closest objects in $\cal X$, and evaluate the retrieved list by \emph{normalized discounted cumulative gain} at position 10 ( NDCG$_{10}$), defined as
\begin{eqnarray*}
{\rm NDCG}_{10} & = &  \frac{\rm DCG_{10}} {\rm Ideal\ DCG_{10}} \\
{\rm Ideal\ DCG_{10}} & = & \sum\limits_{p=1}^{10} \frac{2^{ \sigma^{ (n, i^*_p) } } - 1}{\log_2(1+p)} \\
{\rm DCG}_{10} & = & \sum\limits_{p=1}^{10} \frac{2^{ \sigma^{ (n, i_p) } } - 1}{\log_2(1+p)}
\end{eqnarray*}
where $i_p$ is the $p$th most similar object to $x^{(n)}$ based on the distance measure in test and $i^*_p$ is the $p$th most similar object based on the 
ratings in $\bar{{\cal S}}$. We use NDCG$_{10}$ as our evaluation criterion since it is the most commonly used one when assessing relevance.
\end{enumerate}
The above procedure was repeated for 100 times, resulting in 100 different generative models and data sets. Figure \ref{fig:syn_NDCG} plots the average NDCG$_{10}$ delivered by OR, CO, NCA, and HYB.  The advantage of HYB becomes singificant as the size of the rating data set grows.

\subsection{Real Data}

Our real data set consists of thirty medical images, each corresponding to a distinct CT scan.  Features of each image included semantic annotations given by a radiologist \citep{Rubin08} using a controlled vocabulary and quantitative features such as lesion border sharpness, histogram statistics \citep{Bilello04, Rubin08}, Haar wavelets \citep{Strela99}, and Gabor textures \citep{Zhao04}.
A total of 479 features were extracted from each image, many of which are linearly dependent.  To simplify the computation, we removed those features whose correlations are above 0.95, and normalized the remaining ones.  
This resulted in 60 features which we used in our study.

\begin{figure}[h]
\centering
\includegraphics[scale=0.30]{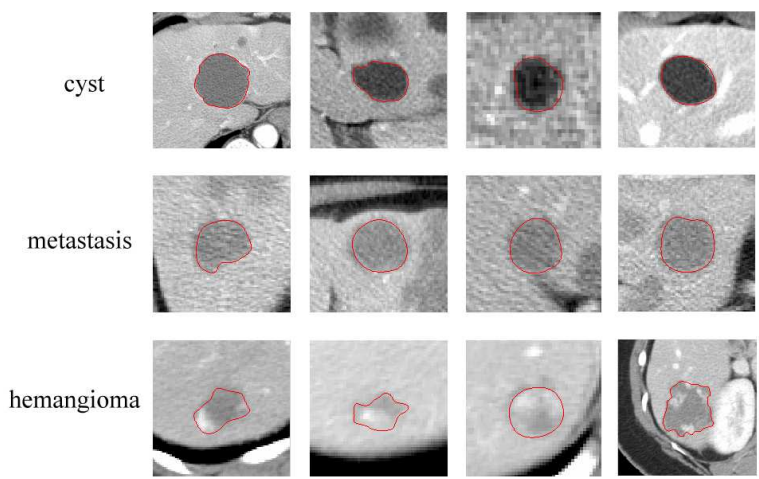}
\caption{Sample images in our data set. Each row of the images corresponds to diagnosis cyst, metastasis, and hemangioma, respectively. The red circles in each image are annotated by a radiologist to indicate the regions of interest.}
\label{fig:example}
\end{figure}

For each pair among the thirty CT scans, we collected two ratings of image similarity  from two different radiologists. Each image was classified with one of three dianoses: cyst, metastasis, or hemangioma. Figure \ref{fig:example} demonstrates some sample images in our data set.

To connect the aforementioned quantities to notation we have introduced, note that the number of features is $K=60$, and the number of classes is $M=3$.
Denote the set of image-feature pairs by ${\cal X}=\{ (o^{(i)},x^{(i)}): 1\leq i \leq 30 \}$, the class label data by ${\cal G} = \{ ( o^{(i)}, x^{(i)}, c^{(i)} ) : 1 \leq i \leq 30 \}$,
and the similarity rating data by ${\cal S} = \{ (o^{(i)}, o^{(j)},x^{(i)},x^{(j)}, \sigma^{(i,j)}) : 1\leq i , j \leq 30, i \neq j\}$.  Tables \ref{tb:ratings} and \ref{tb:classes} provide frequencies with which
different ratings and classes appear in the data set.

\begin{table}[h]
\caption{The distribution of ratings.}
\label{tb:ratings}
\begin{center}
\begin{small}
\begin{sc}
\begin{tabular}{lc}
\hline
Rating & Frequency \\
\hline
1 (Dissimilar) & 58.6\% \\
2 (Neutral) & 16.2\% \\
3 (Similar) & 25.2\% \\
\hline
\end{tabular}
\end{sc}
\end{small}
\end{center}
\end{table}

\begin{table}[h]
\caption{The distribution of classes.}
\label{tb:classes}
\vskip 0.15in
\begin{center}
\begin{small}
\begin{sc}
\begin{tabular}{cc}
\hline
Class & Frequency \\
\hline
Cyst & 44\% \\
Metastasis & 33\% \\
Hemangioma & 23\% \\
\hline
\end{tabular}
\end{sc}
\end{small}
\end{center}
\vskip -0.1in
\end{table}

Since the data points are not very abundant in this case, we use leave-one-out cross-validation to evaluate the performance. More specifically, for $n=1,2,\ldots, 30$, we do the following:
\begin{enumerate}
	\item Let ${\cal X}_{-n} = {\cal X} \setminus (o^{(n)},x^{(n)})$.
	\item Let ${\cal G}_{-n} = {\cal G} \setminus (o^{(n)},x^{(n)}, c^{(n)})$.
	\item Let ${\cal S}_{-n} = {\cal S} \setminus \{(o^{(i)},o^{(j)}, x^{(i)}, x^{(j)}, \sigma^{(i,j)}) : i=n \ {\rm or}\ j=n \}$
	\item Apply the four methods OR, CO, NCA, and HYB on $({\cal X}_{-n}, {\cal G}_{-n}, {\cal S}_{-n})$.
	\item Use each of the resulting distance measures to retrieve the top 10 images from ${\cal X}_{-n}$ that are closest to $x^{(n)}$.
	\item Evaluate the NDCG$_{10}$ of the retrieved lists.
\end{enumerate}

\begin{figure}[htp]
\centering
\includegraphics[scale=0.7]{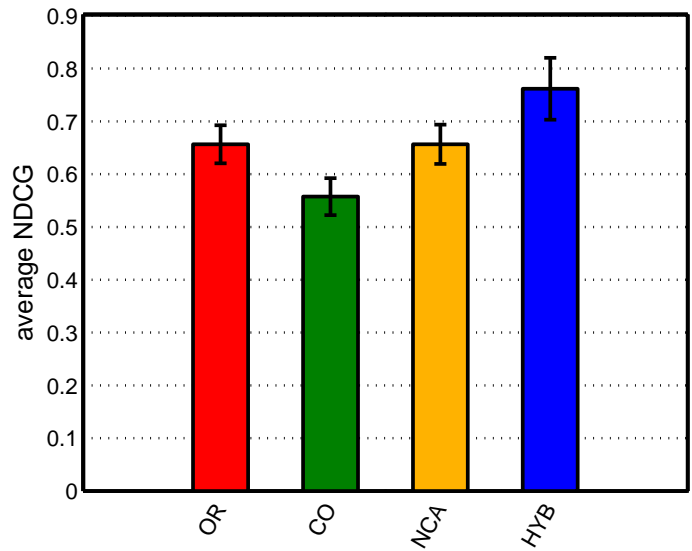}
\caption{The average NDCG$_{10}$ delivered by OR, CO, NCA, and HYB for the medical image data set. For statistical interpretation, we also give the error bars (one standard deviation) in the plots. }
\label{fig:real_NDCG}
\end{figure}

Figure \ref{fig:real_NDCG} plots the average NDCG$_{10}$ delivered by OR, CO, NCA, and HYB. As we can see, HYB leads the other methods by a significant margin of more than 8 percent (0.75 vs. NCA's 0.67).

\section{Conclusion}

We have presented a hybrid method that learns a distance measure by fusing similarity ratings and class labels. This approach consists of two elements, including an algorithm that learns the class probability conditioned on feature through label data, and another algorithm that fits model coefficients so that the resulting distance measure is consistent with similarity ratings.
In our implementation, NCA and CO are chosen for these two elements, respectively. We tried the algorithm on synthetic data as well as a data set collected for the purpose of developing a medical image retrieval system, and demonstrated that it provides substantial gains over various methods that learn distance metrics exclusively from class or similarity data.

As a parting thought, it is worth mentioning that our hybrid method combines elements of generative and discriminative learning.  
There has been a growing literature that explores such combinations \citep{Jaakkola98, Raina03, Kao09} and it would be interesting to explore the relationship
of our hybrid method to other work on this broad topic.

\section*{Appendix: Implementation Details}

\subsection*{$L_1$-regularized NCA}

In our implementation, we randomly partition class label data set $\cal G$ into a training set ${\cal G}_{\rm t}$ and a validation set ${\cal G}_{\rm v}$, whose sizes are roughly $70\%$ and $30\%$ of $\cal G$, respectively. For each $\lambda \in \{ 1, 2, 4, 8, 16\}$, we solve
$$ \max_{r\geq 0} \sum_{(x,c) \in {\cal G}_{\rm t}} \log P\big(c | x, {\cal G}_{\rm t} \setminus (x,c) \big) - \lambda \| r\|_1 $$
by projected gradient ascent. We then compute the log-likelihood of the validation set, given by
$$ \sum_{(x,c) \in {\cal G}_{\rm v}} \log P\big(c | x, {\cal G}_{\rm t} \big),$$
and select the value of $\lambda$ that results in the highest log-likelihood. The resulting value of $\lambda$ is subsequently applied as the regularization parameter when we solve for $r$ with the complete training set $\cal G$. The range of $\lambda$ is determined through trial and error and chosen so that in our experiments the optima rarely took on extreme values.

\subsection*{Sampling Generative Model}
We take $K=20$, $J=20$, and $M=3$ for the synthetic data experiment. Algorithm \ref{alg:sample_gen} is the procedure we use to sample the generative models. Here we set $p(x|c)$ and $p(z|c)$ as mixtures of Gaussian distributions.
This procedure was repeated 100 times to produce 100 generative models.

\begin{algorithm}[h]
   \caption{Sample Generative Model}
   \label{alg:sample_gen}
\begin{algorithmic}
	\FOR{$m=1$ {\bfseries to} $M$}
		\STATE Sample $\alpha_m \sim U[0.5,1.5]$
		\FOR{$i=1$ {\bfseries to} $5$}
			\STATE Sample $\beta_i \sim U[0.5,1.5]$
			\STATE Sample $\mu_{i} \sim {\cal N}(0,I_K)$
			\STATE Sample a matrix $\Sigma_{i} \in \mathbb{R}^{K\times K}$ so that each of its entries is drawn iid from ${\cal N}(0,1/K)$
		\ENDFOR
		\STATE $p(x|m) := \sum\limits_{i=1}^5 \frac{ \beta_i }{\sum_{i'} \beta_{i'} } {\cal N}(x|\mu_{i}, \Sigma^\top_{i} \Sigma_{i}) $
		\FOR{$i=1$ {\bfseries to} $2$}
			\STATE Sample $\gamma_i \sim U[0.5,1.5]$
			\STATE Sample $\phi_{i} \sim {\cal N}(0,I_J)$
			\STATE Sample a matrix $\Omega_{i} \in \mathbb{R}^{J\times J}$ so that each of its entries is drawn iid from ${\cal N}(0,1/J)$
		\ENDFOR
		\STATE $p(z|m) := \sum\limits_{i=1}^2 \frac{ \gamma_i }{\sum_{i'} \gamma_{i'} } {\cal N}(z|\phi_{i}, \Omega^\top_{i} \Omega_{i}) $
   \ENDFOR
	\STATE $P(m) := \frac{ \alpha_m }{\sum_{m'} \alpha_{m'} }, \quad m=1,2,\ldots, M$
	\STATE Sample $r_k \sim \mbox{Exp}(1), \quad k=1,2,\ldots, K$
	\STATE Sample $r^\perp_j \sim \mbox{Exp}(0.2), \quad j=1,2,\ldots, J$
\end{algorithmic}
\end{algorithm}

\bibliography{HDML-ref-01-29-2011}

\end{document}